\documentclass[12pt]{article}

\usepackage{amsmath}
\usepackage{amssymb}
\usepackage{amsthm}
\usepackage{epsfig}
\usepackage{array}
\usepackage{multirow}
\usepackage{xy}
\usepackage{booktabs}
\usepackage[top=1in, bottom=1in,
left=1in,right=1in, headheight=0.0in]{geometry}

\usepackage{graphicx}
\usepackage{subfigure}

\newcommand{\bea}{\begin{eqnarray}}
\newcommand{\eea}{\end{eqnarray}}

\newcommand{\si}{{\sigma}}
\newcommand{\et}{{et al.}}

\newcommand{\lam}{ {\lambda} }
\newcommand{\ta}{{\theta}}
\newcommand{\bt}{ {\beta} }

\newcommand{\gm}{{\gamma}}

\newcommand{\be}{\begin{equation}}
\newcommand{\ee}{\end{equation}}
\newcommand{\mx}{\mathbf{x}}
\newcommand{\my}{\mathbf{y}}

\newcommand{\mf}{\mathbf}

\newcommand{\pa}{\partial}

\newtheorem{thm}{Theorem} [section]
\newtheorem{lemma}{Lemma} [section]

\newenvironment{prf}[1][Proof]{\begin{trivlist}
\item[\hskip \labelsep {\bfseries #1}]}{\end{trivlist}}

\begin{document}

\begin{center}
{\Large\bf Efficient Regularized Regression for Variable Selection with $L_0$ Penalty}
\end{center}

\begin{center}
Zhenqiu Liu$^1$, Gang Li$^2$

\end{center}

\begin{center}
 $^1$Samuel Oschin Comprehensive Cancer Institute, Cedars-Sinai Medical Center, Los Angeles, CA 90048, USA.  $^2$Department of Biostatistics, School of Public Health, University of California at Los Angeles,
Los Angeles, CA 90095-1772, U.S.A.

\end{center}

\begin{center}
\large{\bf Abstract}
\end{center}

Variable (feature, gene, model, which we use interchangeably) selections for regression with  high-dimensional  BIGDATA have  found many applications in bioinformatics, computational biology, image processing, and engineering. One appealing approach is the $L_0$ regularized regression which penalizes the number of nonzero features in the model directly. $L_0$ is known as the most essential sparsity measure and has nice theoretical properties, while the popular $L_1$ regularization is only a best convex relaxation of $L_0$. Therefore, it is natural to expect that $L_0$ regularized regression performs better than LASSO. However, it is  well-known that $L_0$ optimization is NP-hard and computationally challenging. Instead of solving the $L_0$ problems directly, most publications so far have tried to solve  an approximation problem that closely resembles $L_0$ regularization.

In this paper, we propose an efficient EM algorithm  ($L_0$EM) that directly solves the $L_0$  optimization problem.  $L_0$EM is efficient with high dimensional data. It also provides a natural solution to all $L_p$ $p\in [0,2]$ problems, including LASSO with $p=1$,  elastic net with $p \in [1, 2]$, and the combination of $L_1$ and $L_0$ with $p \in  (0, 1]$. The regularized parameter $\lambda$  can be either determined through cross-validation  or AIC and BIC. Theoretical properties of the $L_0$-regularized estimator are given under mild conditions that permit the number of variables to be much larger than the sample size. We demonstrate our methods through simulation and high-dimensional genomic data. The results indicate that $L_0$ has better performance than LASSO and $L_0$ with AIC or BIC has similar performance as computationally intensive cross-validation.    The proposed algorithms are efficient in identifying the non-zero variables with less-bias and selecting biologically important genes and pathways with high dimensional BIGDATA.

\section{Introduction}

Variable selection with regularized regression has been one of the hot topics in machine learning and statistics.  Regularized regressions identify  outcome associated features and estimate nonzero parameters simultaneously,  and are particularly useful for high-dimensional BIGDATA with small sample sizes.  In many real applications, such as  bioinformatics, image and signaling processing, and engineering, a large number of features are measured, but only a small number of features are associated with the dependent variables. Including irrelevant variables in the model will lead to overfitting and deteriorate the prediction performance.  Therefore, different regularized regression methods have been proposed for variable selection and model construction. $L_0$ regularized regressions, which  directly penalize the number of non-zero  parameters, are the most essential sparsity measure.  Several popular information criteria, including Akaike information criterion (AIC) (Akaike 1974), Bayesian information criterion (BIC) (Schwarz 1978), and risk inflation  criteria (RIC) (Foster and George 1994),  are based on $L_0$ penalty and have been used extensively for variable selections.   However, solving a general $L_0$  regularized optimization is NP hard and  computational challenging. Exhaustive  search with AIC  or BIC over all possible combinations of features is computationally infeasible with high-dimensional BIGDATA.

Different alternatives have been proposed  for the regularized regression problem. One common approach is to replace $L_0$ by $L_1$. $L_1$ is known as the best convex relaxation of $L_0$.  $L_1$ regularized regression (Tibshirani 1996) is convex and can be solved by an efficient gradient decent algorithm. Minimizing $L_1$ is equivalent to minimizing $L_0$ under certain conditions. However, the estimates of $L_1$ regularized regression are asymptotically biased, and LASSO may not always choose the true model consistently (Zou 2006). Experimental results by  Mancera and Portilla (2006) also  posed additional doubt about the equivalence of minimizing $L_1$ and $L_0$.  Moreover, there were theoretical results (Lin \et, 2010) showing that while $L_1$ regularized regression never outperforms $L_0$ by a constant, in some cases $L_1$ regularized regression performs infinitely worse than $L_0$. Lin \et (2010) also showed that the optimal  $L_1$ solutions are often inferior to $L_0$ solutions found using greedy classic stepwise regression, although solutions with $L_1$ penalty can  be found effectively.
  More recent approaches aimed to reduce bias and overcome discontinuity include SCAD (Fan and Li, 2001), $L_p$ $p \in (0,1]$ regularization (Liu \et, 2010; Mazumder \et, 2011), and MC+ (Zhang, 2010).  Even though there are some effects for solving the $L_0$ regularized optimization problems (Dicker \et, 2012; Lu $\&$ Zhang, 2013),  $L_0$ was either approximated by a continuous smooth function,  or transformed into a  much larger ranking optimization problem. To the best of our knowledge, there is no method that optimizes  $L_0$  directly.

In this paper, we propose an efficient EM algorithm  ($L_0$EM) that directly solves the $L_0$ regularized regression problem.  $L_0$EM effectively deals with $L_0$ optimization by solving  a sequence of  convex $L_2$ optimizations and  is efficient for high dimensional data. It also provides a natural solution to all $L_p$ $p\in [0,2]$ problems, including LASSO with $p=1$,  elastic net with $p \in [1, 2]$ (Zou $\&$ Zhang 2009), and the combination of $L_1$ and $L_0$ with $p \in  (0, 1]$  (Liu $\&$ Wu, 2007).   While the regularized parameter $\lambda$ for LASSO  must be  tuned through cross-validation, which is time-consuming, the  optimal $\lam$ with $L_0$ regularized regression can be pre-determined  with different model selection criteria  such as AIC, BIC and RIC.  We demonstrate our methods through simulation and high-dimensional genomic data. The proposed methods identify the non-zero variables with less-bias and outperform the LASSO method by a large margin. They can also choose the  biologically important genes and pathways effectively.

\section{Methods}
Given a $n\times 1$  dependent variable $\my$, and an $n \times m$  feature matrix $X$, a linear model is defined as
\[ \my = X\ta + \varepsilon,  \]
where n is the number of samples and m is the number of variables and $n \ll m$, $\ta = [\ta_1, \ldots, \ta_m]^t$ are the m parameters to be estimated, and $\varepsilon \sim N (0, \si^2I_n)$ are the random errors with mean $0$ and variance $\si^2$.  Assume only a small subset of $\{\mx_j\}_{j=1}^m$ has nonzero $\ta_j$s. Let $R \subseteq \{1, \ldots, m\}$ be the subset index of relevant variables with $\ta_j \neq 0$, and $O \subseteq \{1, \ldots, m\}$ be the index of irrelevant features with $0$ coefficients, we have $ R \cup O = \{1, 2, \ldots, m\}$, $X_R \cup X_O = X$, and $\ta_R \cup \ta_O = \ta$, where $\ta_O =0$.  The error function for $L_1$ regularized regression is
\be E = \frac{1}{2}||\my - X\ta||^2 +\frac{\lam}{2}||\ta||_0 =\frac{1}{2}\sum_{i=1}^n (y_i -\sum_{j=1}^m \ta_j x_{ij})^2 + \frac{\lam}{2} \sum_{j =1}^m I(\ta_j \neq 0), \label{eq1} \ee
where $||\ta|| _0 =\sum_{j =1}^m I(\ta_j \neq 0) = |R|$ counts the number of nonzero parameters.
One observation is that equation (\ref{eq1}) is equivalent to  the following equation (\ref{eq2}), when reaching the optimal solution.
\be E = \frac{1}{2}||\my - X\ta||^2 + \frac{\lam}{2}||\ta||_0 = \frac{1}{2}||\my -X\ta||^2 + \frac{\lam}{2} \sum_{j \in R} 1 = \frac{1}{2}||\my -X\ta||^2 + \frac{\lam}{2} |R|, \label{eq2} \ee
because $\ta_O$ is a zero vector. Our  $L_0$EM methods will be derived from equation (\ref{eq2}).
We can rewrite equation (\ref{eq2}) as the following two equations:
\begin{align}  E  &= \frac{1}{2}||\my -X\ta||^2 + \frac{\lam}{2} \sum_{j \in R} \frac{\ta_j^2}{\eta_j^2} \label{eq3} \\
               \eta & =\ta.  \label{eq4}
\end{align}
Given $\eta_j$, equation (\ref{eq3}) is a  convex quadratic function and can be optimized by taking the first order derivative:
\begin{align} \frac{\pa E}{\pa \ta_R} &= \lam\ta_R\oslash \eta_R^2 - X_R^t(\my -X\ta) = 0, \label{eq5} \\
  \textrm{ and } \;\; \frac{\pa E}{\pa \ta_O} & = \mf{0}, \;\; \textrm{ as } \;\; \ta_O =\eta_O= \mf{0},  \label{eq6}      \end{align}
where $\oslash$ indicates element-wise division.  Rewriting (\ref{eq5}) and (\ref{eq6}), we have

\begin{align}
  \lam \ta_R -\eta_R^2\odot X_R^t (\my - X\ta) = 0, \label{eq7}  \\
  \lam \ta_O - \eta_O^2\odot X_O^t(\my - X \ta) = 0, \;\; \forall \;\; \lam > 0 \label{eq8}
\end{align}
where $\odot$ is element-wise multiplication, $\eta_R^2 \odot X_R^t = [\eta^2_R \odot \mx_{1R}^t, \ldots, \eta^2_R\odot \mx_{nR}^t ]$, and $\eta_O^2 \odot X_O^t = [\eta^2_O \odot \mx_{1O}^t, \ldots, \eta^2_O\odot \mx_{nO}^t ] = \mf{0}$.
Let $ X_{\eta}^t = \eta^2\odot X^t$ and combining equations (\ref{eq7}) and (\ref{eq8}) together, we have
\be \eta^2\odot\frac{\pa E}{\pa \ta}=\lam \ta -\eta^2\odot X^t(\my - X\ta) = \lam \ta  - X_{\eta}^t(\my - X\ta) = 0. \label{eq9} \ee
Solving Equation (\ref{eq9}),  we have the following explicit solution.
\begin{align}
      \ta &= (X_{\eta}^tX + \lam I) ^{-1}X_{\eta}^t\my  \label{eq10} \\
      \eta &= \ta, \label{eq11}
\end{align}
where equation (\ref{eq10}) can be considered as the M-step of the EM algorithm maximizing $-E$,  and equation (\ref{eq11}) can be regarded as the E-step with $E(\eta)=\ta$.  Equations (\ref{eq10}) and (\ref{eq11}) together can also be treated as a fixed point iteration method in nonlinear optimization.

%\noindent\textbf{Theorem 1:}
\begin{thm}
\label{theorem1}
Given an input matrix $X$, output matrix $\my$, and initialized solution $\ta^0$, the nonlinear system determined by equations (\ref{eq10}) and (\ref{eq11}) will converge to a unique solution, as long as the regularized parameter $0 < \lam < ||X_\ta^tX||$, and the estimated solution is closer to the true solution after each iterative EM step.
\end{thm}

\noindent\textbf{Proof:} Equations (\ref{eq10}) and (\ref{eq11}) are the same as:
\be \ta = (X_{\ta}^tX + \lam I) ^{-1}X_{\ta}^t\my =(\ta^2\odot X^tX + \lam I)^{-1}(\ta^2\odot X^t)\my \label{eq12} \ee First, $ G(\ta) = (\ta^2\odot X^tX + \lam I)^{-1}(\ta^2\odot X^t)\my$ is  Lipschitz continuous for $\ta \in R^m$, and
\begin{align} \frac{\pa G(\ta)} { \pa \ta} & =(\ta^2\odot X^tX + \lam I)^{-2}[(\ta^2\odot X^tX + \lam I)(2\ta\odot X^t)\my - 2\ta\odot X^tX (\ta^2\odot X^t)\my] \nonumber \\
& = (\ta^2\odot X^tX + \lam I)^{-2}[2\lam\ta\odot X^t \my] = 2\lam (\ta^2\odot X^tX  + \lam I) ^{-1}\mf{1}_m \nonumber \\
& = 2\lam(X_{\ta}^tX + \lam I)^{-1}\mf{1}_m,  \label{eq13}
\end{align}
where $I$ is the identity matrix and $\mf{1}_m=[1, \ldots, 1]^t$ is a $m$-dimensional vector of $1$s, and we substitute equation (\ref{eq12}) into equation (\ref{eq13}) to get the result.

Because $\lam  < ||X_{\ta}^tX||$,  it is clear from equation (\ref{eq13}) that
\[ \left|\frac{\pa G(\ta)}{\pa \ta_j}\right|  = \frac{2\lam}{||X_{\ta}^tX|| + \lam}  < \frac{2\lam}{\lam + \lam} = 1,\]
$\forall$ $j =1,\ldots, m$. Therefore, there is a Lipschitz constant
\[ \gm = \left|\left|\frac{\pa G(\ta)}{\pa \ta}\right |\right|_{\infty}  =\max_j \{\left|\frac{\pa G(\ta)}{\pa \ta_j}\right|\}  < 1. \]
Now given the initial value for equations (\ref{eq10}) and (\ref{eq11}) $\eta = \ta^0 \in R^m$, the sequence $\{\ta^{r}\}$ remains  bounded because $\forall$ $i =1,\ldots, r$,
\begin{align*}  ||\ta^{i+1} -\ta^i||_{\infty}&=||G(\ta^i) - G(\ta^{i-1})||_{\infty}\simeq ||\frac{\pa G(\ta)}{\pa \ta}(\ta^{i} -\ta^{i-1})||_{\infty} \le \left|\left|\frac{\pa G(\ta)}{\pa \ta}\right|\right|_{\infty}||\ta^{i} -\ta^{i-1}||_{\infty}\nonumber \\
& = \gm ||\ta^{i} - \ta^{i-1}||_{\infty}\le \ldots \le \gm^i||\ta^1 -\ta^0||_{\infty}.
\end{align*}
and  therefore
\begin{align*}
||\ta^r - \ta^0||_{\infty} &= ||\sum_{i=0}^{r-1}(\ta^{i+1}-\ta^i)||_{\infty} \le ||\ta^{1} -\ta^0||_{\infty}\sum_{i=0}^{r-1} \gm^i \\
& \le \frac{||\ta^1-\ta^0||_{\infty}}{(1-\gm)}.
\end{align*}
Now $\forall$ $r, k \ge 0$,
\begin{align*}
||\ta^{r+k} - \ta^r||_{\infty} &=||G(\ta^{r+k-1}) - G(\ta^{r-1}||_{\infty} \le \gm ||\ta^{r+k-1}- \ta^{r-1}||_{\infty} \\
                               & \le \gm ||G(\ta^{r+k-2}) - G(\ta^{r-2})||_{\infty} \le \gm^2||\ta^{r+k-2}- \ta^{r-2}||_{\infty} \\
                               & \le\ldots \le \gm^r||\ta^k -\ta^0||_{\infty} \le \frac{\gm^r||\ta^1-\ta^0||_{\infty}}{1-\gm}.
                               \end{align*}
Hence, \[ \lim_{r, k \rightarrow \infty}||\ta^{r+k} - \ta^r||_{\infty}  = 0, \]
and therefore $\{\ta^r\}$ is a Cauchy sequence that has a limit solution $\ta^*$.

Next the uniqueness of the solution is easy to show. Assuming there were two solutions $\ta^*$ and $\ta^{\diamond}$, then
\be ||\ta^{*} - \ta^{\diamond}||_{\infty} = ||G(\ta^{*}) - G(\ta^{\diamond})||_{\infty} \le \gm |\ta^{*} -\ta^{\diamond}||_{\infty}. \label{eq14}\ee
Since $\gm < 1$, equation (\ref{eq14}) can only hold, if $||\ta^{*} -\ta^{\diamond}||_{\infty} = 0$. i.e. $\ta^{*} = \ta^{\diamond}$, so the solution of the EM algorithm is unique.

Finally, the EM algorithm will be closer to the true solution at each step, because
\[ ||\ta^{r+1} - \ta^{*}||_{\infty} =||G(\ta^{r}) - G(\ta^{*}||_{\infty} \le \gm||\ta^{r} -\ta^*||_{\infty}. \]

%\noindent \textbf{Lemma 1:}
\begin{lemma}
\label{lemma1}
Assuming that relevant features are independent, i. e. $\mx_i^t\mx_j = 0$, $\forall$ $i\neq j$  $\&$ $ i, j \in \{1, \ldots, m\}$, then the maximal regularized parameter $\lam$ can be determined by  \[ \lam_{\max} = \max\left\{\frac{(\mx_j^t\my)^2}{4\mx_j^t\mx_j}\right\}_{j=1}^m\]
\end{lemma}

\noindent\textbf{Proof:}  For each feature $\mx_j$ and corresponding coefficient $\ta_j$, equations (\ref{eq9}) and (\ref{eq11}) can be rewritten as
\begin{align*}
 \eta_j^2\frac{\pa E(\ta)}{\pa \ta_j} &= \lam\ta_j - \eta_j^2\mx_j^t (\my - X\ta) = 0  \\
  \eta_j &= \ta_j, \;\;\; \forall \;\; j \in \{1, \ldots, m\}. \end{align*}
  The above two equations are the same as:
  \be \lam\ta_j - \ta_j^2\mx_j^t (\my - X\ta) = 0. \label{eq15}\ee
  If $\ta_j =0$, then any $\lam >0$ will satisfy equation (\ref{eq15}). On the other hand, if
  $\ta_j \neq 0$, because $\mx_i^t\mx_j = 0$, equation (\ref{eq15}) becomes the following quadratic
  equation:
  \be (\mx_j^t\mx_j) \ta_j^2 - (\mx_j^t\my)\ta_j + \lam = 0. \label{eq16} \ee
One necessary condition for equation (\ref{eq16}) to have a solution is:
\[ (\mx_j^t\my)^2 - 4(\mx_j^t\mx)\lam \ge 0,  \;\;\; \Rightarrow \;\; \lam \le \frac{(\mx_j^t\my)^2}{4(\mx_j^t\mx_j)} \]
Therefore the maximal $\lam$ is \[ \lam_{\max} = \max\left\{\frac{(\mx_j^t\my)^2}{4\mx_j^t\mx_j}\right\}_{j=1}^m.\]
If $\lam > \lam_{\max}$,  equation (\ref{eq15}) holds only if all $\ta_j = 0$, $\forall$ $j =1, \ldots, m$.

Both Theorem \ref{theorem1} and Lemma 1 \ref{lemma1} provide some useful guidance for implementing the method and choosing the regularized parameter $\lam$. Theorem \ref{theorem1} shows that the EM algorithm always converges to a unique solution, given a certain $\lam$ and initial solution $\ta^0$, and the estimated value is closer to the true solution after each EM iteration. Note that different initial values may still reach different solution, because of the non-convex $L_0$ penalty. Therefore, it is critical to choose a  good initial value. Our experiences with the method indicate that initializing with the estimates from $L_2$ based ridge regression will usually lead to quick converge and super performance.  The EM algorithm is as follows.

\begin{tabular}{l}
\textbf{The $L_0$EM Algorithm:}\\\hline
Given a $0<\lam\le \lam_{\max}$, small numbers $\epsilon$ and $\varepsilon$,\\
 and training data $\{X, \my\}$,\\
Initializing $\ta =(X^tX +\lam I)^{-1}X^t\my$, \\
While 1,\\
 \;\;\;\;\; E-step:   $ \mf{\eta} = \ta $  \\
\;\;\;\;\;  M-step: $X_{\eta}^t = \eta^2\odot X^t = [\eta^2\odot\mx_1 ,\ldots, \eta^2\odot \mx^t_n]$\\
\;\;\;\;\;\;\;\;\;\;\;\;\;\;\;\text{ }\text{ } \text{ }$ \ta = (X_{\eta}^tX + \lam I) ^{-1}X_{\eta}^t\my $\\
\;\;\;\;\;\;if $||\ta -\eta|| < \varepsilon$, Break;  End\\
End \\
 $\ta(|\ta| <\epsilon) = 0$. \\ \hline\\
\end{tabular}

\noindent Similar procedures can be extended to general $L_p$, $p \in [0, 2]$ without much difficulty. $L_p$  based EM algorithm $L_p$EM  is reported in Appendix.

\noindent\textbf{Consistency and Oracle Property:}
Let $\theta_0$ be the true parameter value. The following conditions will be used later for theoretical properties of the $L_0$-regularized estimator of $\theta_0$.

\noindent {\it CONDITIONS}
\begin{itemize}
\item[(C1)] $ln(m)=o(n)$ as $n \to \infty$.

\item[(C2)] There exists a constant $K>0$ such that
$\lambda_{max}(\frac{ X^tX}{n})\le K < \infty$ for  large $n$, where for any matrix $B$, $\lambda_{max}(B)$ denotes the largest eigenvalue of $B$.

\item[(C3)] $\frac{\max_j || \mx_j||}{\sqrt{n}} =O(\sqrt{ln(mn)})$ or $O(1)$ as $n,m\to \infty$.

\item[(C4)] There exists a constant $c>0$ such that $\frac{\min_j ||\mx_j||^2}{n}  \ge c>0 $ for large $n,m$.

\item[(C5)] $\mu(X) \equiv \max_{1\le i<j\le m} \frac{|\mx_i^t \mx_j | }{||\mx_i||\cdot ||\mx_j||}=O(\sqrt{\frac{ln(m)}{n}})$.

\item[(C6)] $||\theta_0||_0=O(1)$.

\end{itemize}
The above conditions are very mild. Condition (C1) trivially holds for $m\le n$ and for $m>n$. In particular, (C1) is satisfied even for ultra-high dimensional case such as $m=exp(n^{\alpha})$ for $0<\alpha<1$. (C2) is a standard condition for linear regression. Chi (2013, Section 3.2) gives examples satisfying(C3)-(C4). For example, (C3) and (C4) trivially hold if $||\mx_i = \sqrt{n}$ for all $j=1,\ldots,m$. (C5) is referred to as the coherence condition under which the covariates are not highly colinear; see Bunea et al. (2007), Candes and Plan (2009), and Chi (2013). (C6) implies that the model is sparse.

The following theorem is a direct consequence of Chi (2013).
\begin{thm}[{\em Consistency}]
\label{theorem2}
Assume that conditions (C1)-(C6) hold.
Let $n(\nu)=(1-\nu)[1 + 1/\mu(X)]$ for some $0<\nu<1$.
For any $0<q<\frac{1}{2}$, let
$\lambda =\frac{3ln(m/q)}{\nu [1+\mu(X) ]} \frac{\max_j || \mx_j||^2}{\min_j ||\mx_j||^2}$
and
\begin{equation}
\nonumber
\hat\theta = arg \min_{||\theta||_0 \le n(\nu)} E_n(\theta),
\end{equation}
 Then, with probability tending to 1,
\begin{align}
\label{equation17}
|| \hat{\theta} - \theta_0|| = O_p(\sqrt{\frac{\ln(nm)}{n}})
\end{align}
\end{thm}

\begin{prf} Note that the normal linear model  in this paper is a special case of
the exponential model of Chi (2009): $p_t(y) =\exp(ty -\Lambda(t))$
with $t=\frac{\mx_i^t \theta}{\sigma^2}$ and $\Lambda(t) = \frac{\sigma^2 t^2}{2}$.
Then, (\ref{equation17}) follows immediately from Theorem 3.1 of Chi (2009).
\end{prf}

\noindent\textbf{Model Recovery:} Next we show that $L_0$-regularized regression recovers the true model under mild conditions.
\begin{thm}[{\em Oracle Property}]
\label{theorem3}
Assume that conditions (C1)-(C6) hold.
Let
$
A = \{1 \leq j \leq m: \theta_{0j} \neq 0\},
$ and
$
A^c = \{1, 2, \ldots, m\} \backslash A.
$
Then, the minimizer $\hat{\theta}$ in Theorem \ref{theorem2} must satisfy $\hat{\theta}_j = 0$ for $j \in A^c$.
\end{thm}

\begin{prf} Let $\alpha_n=\sqrt{\frac{\ln(nm)}{n}}$.
For any $\theta$ such that $||\theta - \theta_0|| < C \alpha_n$ for some constant $C>0$ and $\displaystyle \sum_{j \in A^c} I(\theta_j \neq 0) \ge 1$,  let
\begin{align*}
   \tilde{\theta}_j =     \left\{ \begin{array}{rcl}
         \theta_j & \mbox{if} & j \in A \\
 	 0  & \mbox{if} & j \in A^c
                \end{array}\right.
\end{align*}
Then,
\begin{align*}
& E_n(\theta) - E_n(\tilde{\theta}) \\
& = \frac{1}{2n} (\theta - \tilde{\theta})^TX^TX(\theta-\tilde{\theta}) - \frac{1}{n}(\theta-\tilde{\theta})^TX^T(y-X\tilde{\theta}) + \frac{\lambda}{2} (||\theta||_0-||\tilde{\theta}||_0) \\
& = \frac{1}{2n}(\theta-\tilde{\theta})^TX^TX(\theta-\tilde{\theta}) - \frac{1}{n}(\theta-\tilde{\theta})^TX^T(X\theta_0+\epsilon - X \tilde{\theta}) + \frac{\lambda}{2}(||\theta||_0 - ||\tilde{\theta}||_0) \\
& = \frac{1}{2} (\theta-\tilde{\theta})^T \left(\frac{X^TX}{n}\right)(\theta-\tilde{\theta}) - (\theta-\tilde{\theta})^T \left(\frac{X^TX}{n}\right)(\theta_0-\tilde{\theta}) + \\
& +\frac{1}{\sqrt{n}} (\theta-\tilde{\theta})^T \cdot \frac{1}{\sqrt{n}}  X^T\epsilon + \frac{\lambda}{2} (||\theta||_0-||\tilde{\theta}||_0) \\
& = I_1 + I_2 + I_3 + I_4
\end{align*}
Because $||\tilde{\theta}-\theta_0|| \leq ||\theta-\theta_0||$, we have $\theta-\tilde{\theta} = O(\alpha_n)$.
Thus,
$I_1 = O(\alpha_n^2)$ and
$I_2 = O(\alpha_n^2)$.
Moreover,
\begin{align*}
\bigg| \bigg| \frac{1}{\sqrt{n}} \epsilon^tX \bigg| \bigg|
= O_p(\sqrt{k\sigma^2}), \hspace{.2in}\mbox{ as } n \to \infty
\end{align*}
where $k=rank(X)\le n$.
Hence,
\begin{align*}
|I_3| \leq \frac{1}{\sqrt{n}} ||\theta-\tilde{\theta}|| \cdot \bigg| \bigg| \frac{1}{\sqrt{n}} X^T\epsilon \bigg| \bigg| = O(\alpha_n)\cdot O_p(\sqrt{k/n}) = O_p(\alpha_n).
\end{align*}
Furthermore,
\begin{align*}
I_4 & = \frac{\lambda}{2} ( ||\theta||_0 - ||\tilde{\theta}||_0) \\
& = \frac{\lambda}{2} \sum_{j=1}^m [I(\theta_j \neq 0) - I(\tilde{\theta}_j \neq 0)] \\
& = \frac{\lambda}{2} \left[\sum_{j \in A} 0 \right] + \frac{\lambda}{2} \sum_{j \in A^c} [I(\theta_j \neq 0) - 0] \\
& = \frac{\lambda}{2} \sum_{j \in A^c} I(\theta_j \neq 0) \geq \frac{\lambda}{2} \cdot 1 > 0.
\end{align*}
By conditions (C3)-C(5), $\lambda = O(ln(m) \cdot ln(nm))$. Therefore, the first three terms $I_1$, $I_2$ and $I_3$ are dominated by $\lambda$ in probability as $n\to\infty$.
Therefore, with probability tending to 1,
\begin{align}
E_n(\theta) - E_n(\tilde{\theta}) > 0.
\end{align}
This completes the proof of Theorem \ref{theorem3}.
\end{prf}

\noindent\textbf{Determination of $\lam$:} The regularized $\lam$ determines the sparsity of the model.  The standard approach for choosing $\lam$ is cross-validation and the optimal $\lam$  is determined by  the minimal mean squared error (MSE) of the test data  ($ MSE = \sum(y_i -\hat{y_i})^2/n$). One could also adapt the  stability selection (SS) approach for $\lam$ determination (Liu \et 2010; Meinshausen, 2010). It chooses  the smallest $\lam$  that minimizes the inconsistences in number of nonzero parameters with cross-validation.  We first calculate the mean and standard deviation (SD) of the number of nonzero parameters for each $\lam$, and then find the smallest $\lam$ with 0 SD, where 0 SD indicates that all models in k-fold cross validation has the same number of nonzero estimates.  Our experiences indicate that the larger $\lam$ chosen from both minimal MSE and stability selection ($\lam = \max \{\lam_{mse}, \lam_{ss}\}$) has the best performance. Choosing optimal $\lam$ from cross-validation is computationally intensive and time consuming. Fortunately, unlike LASSO,  identifying the optimal $\lam$ for $L_0$ does not  require to use cross validation.  The optimal $\lam_{opt}$  can be determined by variable selection criteria. The optimal $\lam_{opt}$ can be directly picked using  AIC, BIC, or RIC criteria with $\lam_{opt} = 2$, $\log n$, or $2\log m$, respectively. Each of these criteria is known to be optimal under certain conditions.  This is a huge advantage of $L_0$, especially for BIGDATA problems.

\section{Simulations}
To evaluate  the performance  of  $L_0$ and $L_1$ regulation, we assume a linear model $\my = X\ta +\varepsilon$, where the input matrix $X$ is from Gaussian distribution with mean $\mu = \mf{0}$ and different covariance structures $\Sigma$, where $\Sigma(i, j) = r^{|i-j|}$ with $r = 0, 0.3, 0.6, 0.8$ respectively. The true model is $\my = 2\mx_1 - 3\mx_2 + 4\mx_ 5 + \varepsilon$ with $\varepsilon \sim N(0, 1)$. Therefore, only three features are associated with output $\my$, and the rest of the $\ta_i$s are zero.  In our first simulation,  we first compare $L_0$ and $L_1$ regularized regression with a relative small number of features $m = 50$ and a sample size of n =100.  Five-fold cross validation is used to determine the optimal $\lam$ and compare the model performance. We seek to fit the regularized regression models over a range of regularization parameters $\lam$. Each $\lam$ is chosen from $\lam_{\min} = 1e-4$, to $\lam_{\max}$ with 100 equally log-spaced intervals, where $\lam_{\max} = \max\{X^t\my\}$ for $L_1$ and $\max\left\{\frac{(\mx_j^t\my)^2}{4\mx_j^t\mx_j}\right\}$ for $L_0$. Lasso function in the statistics toolbox of MATLAB (www.mathworks.com) is used for comparison.  Cross-validation with MSE is implemented nicely in the toolbox. The computational results are reported in Table 1.
\begin{table}[t!]
 \begin{center}
 \begin{tabular}{c|ccc|ccc} \hline
   \multirow{2}{*}{r} & \multicolumn{3}{c|}{$L_0$}  & \multicolumn{3}{c}{$L_1$}\\\cmidrule{2-7}
      &$\#$ SF & MSE  &   $||\hat{\ta} -\ta||$ & $\#$ S.F. & MSE & $||\hat{\ta}-\ta||$  \\ \hline
    0& $3.39(\pm 1.1)$  & $1.01 (\pm 0.14)$ & $0.206(\pm 0.12)$ & $14.5(\pm 3.45)$ & $1.19(\pm 0.19)$  & $0.38(\pm 0.1)$\\
    0.3 &$3.37(\pm 0.9)$ & $1.02 (\pm 0.16)$ &$0.23(\pm 0.12)$ & $14.5(\pm 2.91)$ & $1.21(\pm 0.19)$   & $0.41 (\pm 0.19)$  \\
    0.6& $3.49(\pm 1.7)$ & $1.02 (\pm 0.23)$   & $0.23(\pm 0.16)$  & $13.5(\pm 3.0)$ & $1.26(\pm 0.2)$  &$0.54 (\pm 0.15)$   \\
    0.8 &$3.32(\pm 0.9)$ & $1.06 (\pm 0.15)$   & $0.28(\pm 0.21)$  &$11.7(\pm 2.69)$  & $1.3(\pm 0.21)$ & $0.89(\pm 0.25)$ \\\hline
\end{tabular}
\caption{\footnotesize Performance measures for $L_0$ and $L_1$ regularized regression over 100 simulations, where values in the parenthesis are the standard deviations, and  $\#$ SF: number of average selected features; MSE: Average mean squared error; $||\hat{\ta} -\ta||$:  average absolute bias when comparing true and estimated parameters.}
\end{center}
\end{table}
Table 1 shows that $L_0$ outperforms LASSO in all categories by a substantial margin, when using the popular test MSE measure for model selection. In particular, the number of variables selected by $L_0$ are very close  to the true number of variables (3), while LASSO selected more than 11 features on average with different correlation structures (r = 0, 0.3, 0.6, 0.8).  The test MSEs and bias both increase with the growth of correlation among features for both $L_0$ and LASSO, but the test MSE and bias of $L_0$ are substantially lower than these of LASSO. The maximal MSE of $L_0$ is 1.06, while the smallest MSE of $L_1$ is 1.19, and the largest bias of $L_0$ is 0.28, while the smallest bias of LASSO is 0.38. In addition (results are not shown in Table 1), $L_0$ correctly identifies the true model  81, 74, 81,  and 82  times for r = 0, 0.3, 0.6, and 0.8 respectively over 100 simulations, while LASSO never chooses  the correct model. Therefore, compared to $L_0$ regularized regression, LASSO selects more features than necessary and  has larger bias in parameter estimation. Even  though it is possible to get a correct  model  with LASSO using a larger $\lam$,  the estimated parameters will have a bigger bias and worse predicted MSE.

The same parameter setting  is used for our second simulation, but the regularized parameter $\lam$ is determined by the larger $\lam$ from both minimal MSE and stability selection ($\lam = \max\{\lam_{MSE}, \lam_{SS}\}$).  The
computational results are reported in Table 2.
\begin{table}[thb!]
 \begin{center}
 \begin{tabular}{c|ccc|ccc} \hline
   \multirow{2}{*}{r} & \multicolumn{3}{c|}{$L_0$}  & \multicolumn{3}{c}{$L_1$}\\\cmidrule{2-7}
      &$\#$ SF & MSE  &   $||\hat{\ta} -\ta||$ & $\#$ S.F. & MSE & $||\hat{\ta}-\ta||$  \\ \hline
    0& $3.09(\pm 0.53)$  & $1.04 (\pm 0.15)$ & $0.18(\pm 0.11)$ & $13.3(\pm 4.56)$ & $1.21(\pm 0.17)$  & $0.39(\pm 0.1)$\\
    0.3 &$3.08(\pm 0.54)$ & $1.04 (\pm 0.15)$ &$0.17(\pm 0.07)$ & $14.5(\pm 4.20)$ & $1.22\pm 0.17)$   & $0.42 (\pm 0.19)$  \\
    0.6& $3.10 (\pm 0.46)$ & $1.07 (\pm 0.17)$   & $0.21(\pm 0.10)$  & $13.8(\pm 5.4)$ & $1.27(\pm 0.47)$  &$0.57 (\pm 0.25)$   \\
    0.8 &$3.02(\pm 0.14)$ & $1.04 (\pm 0.14)$   & $0.26(\pm 0.13)$  &$13.4(\pm 4.91)$  & $1.25(\pm 0.21)$ & $0.74(\pm 0.25)$ \\\hline
\end{tabular}
\caption{\footnotesize Performance measures for $L_0$ and $L_1$ regularized regression with $\lam = \max\{\lam_{mse}, \lam_{ss}\}$ over 100 simulations, where values in the parenthesis are the standard deviations, and  $\#$ SF: number of average selected features; MSE: Average mean squared error; $||\hat{\ta} -\ta||$:  average absolute bias when comparing true and estimated parameters.}
\end{center}
\end{table}
Table 2 shows that the average number of associated features is much closer to 3 with sightly larger test MSEs.  The maximal average number of features is 3.1 with $r = 0.6$, reduced from 3.49 with the test MSE only.  In fact, with this combined model selection criteria and 100 simulations, $L_0$EM identified the true model with three nonzero parameters 95, 95, 95, and 97 times respectively (not shown in the table), while LASSO did not choose any correct models. The average bias of the estimates with $L_0$EM  is also reduced. These indicate that the combination of test MSE and stability selection in cross-validation  leads to  better  model selection results  than MSE alone with $L_0$EM.  However, the computational results did not improve much with LASSO.  Over 13 features on average  were selected under different correlation structures, suggesting that LASSO inclines to select more spurious features than necessary. A  much more conservative criteria with larger $\lam$ is required to select the right number of features, which will induce larger MSE and bias, and deteriorate the prediction performance.

\subsection*{Simulation with high- dimensional data}
 Our third simulation deals with high-dimensional data with the number of  samples $n =100$, and the number of features $m = 1000$.  The correlation structure is set to $ r =0, 0.3, 0.6$, and the same model $\my = 2 \mx_1 - 3\mx_2 + 4\mx_5 + \varepsilon $ was used for  evaluating the performance of  $L_0$ and $L_1$. The simulation was repeated 20 times. The computational results are reported in Table 3.
 \begin{table}[thb!]
 \begin{center}
 \begin{tabular}{c|c|ccc} \hline
    & Measures & r = 0  & r = 0.3     & r = 0.6 \\ \hline
   \multirow{4}{*}{$L_0$} & $\#$ SF &  $3(\pm 0)$  & $2.9(\pm 0.47)$ & $2 (\pm 0.73)$ \\
      &         $||\hat{\ta} -\ta||$  &  $0.14 (\pm 0.09)$ & $0.39(\pm 0.63)$ & $1.69(\pm 1.25)$ \\
      &          Test MSE            & $1.14(\pm 0.34) $  &  $1.59 (\pm 1.3)$  & $ 2.8 (\pm 1.72)$ \\
      &         $\#$ True Model  &  20/20      &  15/20 &   5/20 \\ \hline
      \multirow{4}{*}{$L_1$} & $\#$ SF &  $24(\pm 18.4)$  & $31.3(\pm 20.7)$ & $36.7 (\pm 16.5)$ \\
      &         $||\hat{\ta} -\ta||$  &  $0.57 (\pm 0.11)$ & $0.73(\pm 0.13)$ & $1.14(\pm 0.25)$ \\
      &          Test MSE            & $1.50(\pm 0.25) $  &  $1.63 (\pm 0.29)$  & $ 1.92(\pm 0.41)$ \\
      &         $\#$ True Model  &  0/20      &  0/20 &   0/20 \\ \hline

\end{tabular}
\caption{\footnotesize Performance measures for $L_0$ and $L_1$ regularized regression with cross validation and $\lam = \max\{\lam_{mse}, \lam_{ss}\}$ over 20 simulations and the sample size of $n=100$, and $m =1000$, where values in the parenthesis are the standard deviations, and  $\#$ SF: number of average selected features; MSE: Average mean squared error; $||\hat{\ta} -\ta||$:  average absolute bias when comparing true and estimated parameters.}
\end{center}
\end{table}
Table 3 shows that $L_0$ outperforms LASSO by a large margin when correlations among features are low.  When there is no correlation among features, 20 out of 20 simulations  identify the true model with $L_0$, and 15 out of  20 simulations choose the correct model when $r = 0.3$, while LASSO again  chooses more features than necessary and no true model was found  under any  correlation setting.   However, when correlations among features are large with $r =0.6$, the results are mixed.  $L_0$ can still identify 5 out of 20 correct models, but the test MSE and bias of the parameter estimate of $L_0$ are slightly large than those of LASSO. In addition, we notice that $L_0$ is a more sparse model when correlation increases, indicating that $L_0$ tends to choose independent features.  The regularization path of $L_0$   regression is shown in Figure 1.
\begin{figure}[htb!]
\centering
 \includegraphics[scale=0.55]{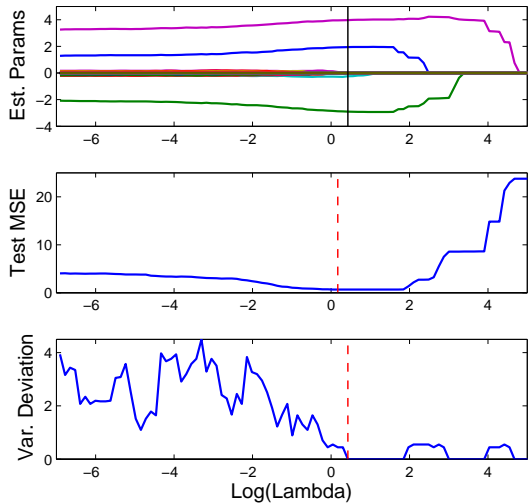}
\caption{\footnotesize  Regularized path for $L_0$ penalized regression with n=100, m =1000, and r = 0.3}
   \end{figure}
   As shown in the top panel of Figure 1,  the three associated features first increase their values when $\lam$ goes larger, and then go to zero when $\lam$ becomes extremely big, while the rest of the irrelevant features all go to zero when $\lam$ increases.  Unlike LASSO, which shrinks all parameters uniformly, $L_0$ will only forces the estimates of irrelevant features go to zero, while keep the estimates of relevant features to their true value. This is the  well-known Oracle property of $L_0$.  For this specific simulation, the three parameters $[\hat{\ta_1}, \hat{\ta _2}, \hat{\ta_5}] = [ 1.85, -2.94, 4.0]$, very close to their true values $[2, -3, 4]$.  The middle  and bottom panels are the test MSE and  the standard deviation of the number of nonzero variables. The optimal $\lam$ is chosen from the the larger $\lam$ with minimal test MSE and stability selection as shown in the vertical lines of Figure 1.
    \subsubsection*{$L_0$ regularized regression without cross validation} Choosing the optimal parameter $\lam_{opt}$ with cross-validation is time consuming, especially with BIGDATA. As we mentioned previously, the optimal $\lam$ can be picked from theory instead of cross validation.  Since  we are dealing with the $n \ll m$  BIGDATA problem, RIC with $\lam_{opt} = 2 \log m$ is penalized too much for such problem.  So computational results with AIC and BIC without cross validation are reported in Table 4.
\begin{table}[t!]
 \begin{center}
 \begin{tabular}{c|c|ccc} \hline
    & Measures & r = 0  & r = 0.3     & r = 0.6 \\ \hline
   \multirow{4}{*}{AIC} & $\#$ SF &  $3.26(\pm 0.54)$  & $3.72(\pm 1.94)$ & $4.8 (\pm 2.77)$ \\
      &         $||\hat{\ta} -\ta||$  &  $0.19 (\pm 0.09)$ & $0.36(\pm 0.58)$ & $1.02(\pm 1.2)$ \\
      &          MSE$^*$            & $0.96(\pm 0.14) $  &  $1.02 (\pm 0.31)$  & $ 1.27 (\pm 0.51)$ \\
      &         $\#$ True Model  &  78/100     &  73/100 &   59/100 \\ \hline
      \multirow{4}{*}{BIC} & $\#$ SF &  $3.0(\pm 0.0)$  & $3.0(\pm 0.38)$ & $2.89 (\pm 0.80)$ \\
      &         $||\hat{\ta} -\ta||$  &  $0.16 (\pm 0.08)$ & $0.45(\pm 0.69)$ & $1.80(\pm 1.20)$ \\
      &          MSE$^*$            & $0.97(\pm 0.15) $  &  $1.29 (\pm 0.81)$  & $ 2.48(\pm 1.17)$ \\
      &         $\#$ True Model  &  100/100      &  94/100 &   53/100 \\ \hline

\end{tabular}
\caption{\footnotesize Performance measures for $L_0$ regularized regression with AIC and BIC  over 100 simulations with $n=100$, and $m =1000$, where values in the parenthesis are the standard deviations, and  $\#$ SF: number of average selected features; MSE$^*$: In-sample average mean squared error; $||\hat{\ta} -\ta||$:  average absolute bias when comparing true and estimated parameters.}
\end{center}
\end{table}
Table 4 shows that $L_0$ regularized regression with AIC and BIC performs very well, when compared with  the results from computationally intensive cross-validation in Table 3.  Without correlation,   BIC identifies the true model ($100\%$), which is the same as cross-validation  in Table 3, and better than AIC's $78\%$. The bias of BIC (0.16) is only slightly higher than that of  cross-validation (0.14), but lower than that of AIC (0.19). Even though MSE$^*$s with AIC and BIC are in-sample mean squared errors, which are  not comparable to the test MSE with cross validation,  larger MSE$^*$ with BIC indicates that BIC is an more stringent criteria than AIC and selects less variables.  With mild correlation ( r = 0.3) and some sacrifices in bias and MSE$^*$ , BIC seems to perform the best in variable selection, since the average number of features selected is exactly 3 and $94\%$  of the simulations recognize the true model, while AIC chooses more features (3.72) than necessary  and  only $73\%$ of the simulations are right on targets. Cross validation is the most tight measure with 2.9 features on average and $75\%$ of the simulations finding the correct model.  When the correlations among the variables are  high (r= 0.6), the results are mixed. Both BIC and AIC correctly identify more than half of the true models, while cross validation only recognizes $25\%$ (5/20) of the model correctly.  Therefore, comparing with the computationally intensive cross validation,  both BIC and AIC perform reasonable well. The computational  results of BIC is comparable to the results of cross validation, while the computational time is only $1/500$ of the time for cross validation, if the free-parameter $\lam_{opt}$ is chosen from 100 candidate $\lam$s with 5-fold cross validation.

\subsection*{Simulations for graphical models}
One important application of $L_0$ regularized regression is to detect  high-order correlation structures, which has numerous real-world applications including gene network analysis.  Given a matrix $X$, letting $\mx_j$ be the $j$th variable, and $X_{-j}$ be the remaining variables, we have $P(\mx_j|X_{-j}) \sim N(X_{-j}\ta, \si^2),$  where the coefficients $\ta$  measures the partial correlations between $\mx_j$ and the rest variables.  Therefore, the high-order structure of X has been determined via a series of $L_1$ regularized regression  for each $\mx_j$ with the remaining variables $X_{-j}$ (Peng \et 2009; Liu $\&$ Ihler, 2011). The collected regression nonzero coefficients are the edges on the graph. The drawback of such approach is computationally intensive, because the regularized parameter $\lam$ for $L_1$  have to be determined through cross validation. For instance, given a matrix $X$ with 100 variables,   to find the optimal $\lam_{opt}$ from 100 candidate $\lam$s with 5-fold cross validation, 500 models need to be evaluated  for each variable $\mx_j$. Therefore a total of $500\times 100 = 50000$ models have to be estimated to detect the dependencies among $X$ with LASSO. It usually takes hours to solve  this problem.  However,  only 100 models are required to identify the same correlation structure with $L_0$ regularized regression and AIC or BIC. Solving such a problem with $L_0$ without cross-validation only takes less than one minute. Finally, negative correlations between genes are difficult to confirm and seemingly less ‘biologically
relevant’ (Lee \et, 2004).  Most national databases are constructed with similarity (dependency) measures. it is straight forward to study only the positive dependency by simply setting $\ta(\ta <0) = 0$ in the EM algorithm.

We simulate two network structures similar to those in Zhang $\&$ Mallick (2013) (i) Band 1 network, where $\Sigma$ is  a covariance matrix with $\si_{ij} = 0.6^{|i-j|}$, so $A =\Sigma^{-1}$ has a  band 1 network structure, and (ii) A more difficult problem for a Band 2 network with weaker correlations, where $A = -\Sigma^{-1}$ with $a_{ij} =\left\{\begin{array}{rl}
0.25     & \textrm{ if } |i-j| = 1,\\
 0.4      &  \textrm{ if } |i-j| =2, \\
  0        & \textrm{ Otherwise. }
  \end{array} \right. $
The sample sizes are n = 50, 100, and 200, respectively and the number of variables is $m = 100$.  $L_0$ regularized regression with AIC and BIC is used to detect the network (correlation) structure. The consistence between the true and predicted structures is measured by the area under the ROC curve (AUC), false discovery (positive) rate (FDR/FPR), and false negative rate (FNR) of edges.
 The computational results are shown in Table 5.
\begin{table}[t!]
 \begin{center}
 \begin{tabular}{c|ccc|ccc} \hline
    & \multicolumn{3}{c|}{AIC}  & \multicolumn{3}{c}{BIC}\\ \hline
    Band 1  & AUC & FDR($\%$)  &   FNR ($\%$) &  AUC & FDR ($\%$) & FNR ($\%$) \\ \hline
    $n=50$& $.95(\pm .01)$  & $.29 (\pm .08)$ & $9.4(\pm 2.6)$ & $.90(\pm .02)$ & $.10(\pm .05)$  & $20(\pm 3.6)$\\
    100 &$.99(\pm .005)$ & $.20 (\pm .06)$ &$1.2(\pm 1.1)$ & $.991(\pm .007)$ & $.03(\pm .03)$   & $1.8 (\pm 1.3)$  \\
    200& $.999(\pm .0003)$ & $.20 (\pm .05)$   & $0(\pm 0)$  & $.9999(\pm .0005)$ & $.01(\pm .01)$  &$.01(\pm .10)$   \\ \hline
    Band 2 & AUC & FPR($\%$)  &   FNR ($\%$) &  AUC & FPR ($\%$) & FNR ($\%$) \\ \hline
      $n =50$  & $.82(\pm .01)$  & $.10 (\pm .05)$ & $36.7(\pm 1.5)$ & $.803(\pm .008)$ & $.02(\pm .02)$  & $ 39.3(\pm1.5)$ \\
      100     & $.84(\pm.01)$ & $.11(\pm .04)$ & $32.7(\pm 1.9)$  &$.83(\pm .01)$ & $.03(\pm .02)$ & $ 34.9(\pm 1.6)$  \\
      200  & $.93(\pm .01)$ & $.11(\pm .04)$ & $14.2 (\pm 2.4)$ & $.82(\pm .01)$ & $.03 (\pm .02)$ & $ 36.7 (\pm 1.8)$ \\ \hline
\end{tabular}
\caption{\footnotesize Performance measures for $L_0$ regularized regression for graphical structure detection over 100 simulations, where values in the parenthesis are the standard deviations.}
\end{center}
\end{table}
Table 5 shows that both AIC and BIC performed well. Both achieved at least 0.90 AUC for Band 1 network and 0.8 AUC for Band 2 network with different sample sizes.   AIC performed slightly better than BIC, especially for  Band 2 network with weak correlations and small sample sizes.  This is reasonable because BIC is a heavier penalty and forces most of the weaker correlations with $a_{ij} = 0.25$ to 0.  In addition, BIC has slightly larger AUCs for Band 1 network with strong correlation $r =0.6$ and larger sample size (n=100, 200). One interesting observation is that  the FDRs of both AIC and BIC are well controlled. The maximal FDRs  of AIC for the Band 1 and  2 networks are $0.29\%$ and $0.2\%$, while the maximal FDRs of BIC are only $0.1\%$, and $0.03\%$ respectively. Controlling false discovery rates is crucial for identifying true associations with high-dimensional data in  bioinformatics. In general, AUC increases  and both FDR and FNR decrease, as the sample sizes become larger, except for Band 2 network with BIC. The performance of BIC is not necessary better with large sample size, since the penalty $\lam$ increases with the sample size.

\section{Real Application}
The purpose of this application is to identify subnetworks and study the biological mechanisms of potential prognostic biomarkers for ovarian cancer with multi-source gene expression data. The ovarian cancer data was downloaded from the KMplot website(www.kmplot.com/ ovar) (Gyorffy \et 2012). They  originally got the data from searching Gene Expression Omnibus (GEO; http://www.ncbi.nlm.nih.gov/geo/) and The Cancer Genome Atlas (TCGA;  http:// cancergenome.nih.gov) with multiple platforms.  All collected datasets have  raw gene expression data,  survival information, and at least 20 patients available. They merged the datasets across different platforms carefully. The final data has 1287 patients samples, and 22277 probe sets representing 13435 common genes. We identified 112 top genes that are associated with patient survival times using univariate COX Regression.  We constructed a co-expression network from the 112 genes with $L_0$ regularized regression and identified biologically meaningful subnetworks (modules) associated with patient survival.  Network  is constructed with positive correlation only and BIC. The computational time for constructing such network is less than 2 seconds.  One survival associated subnetwork we identified is given in Figure 2.
\begin{figure}[htb!]
\centering
 \includegraphics[scale=0.5]{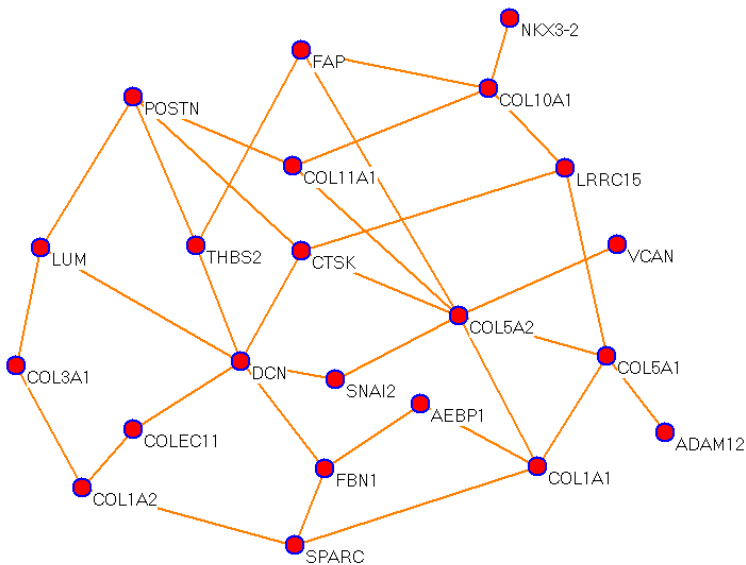}
\caption{\footnotesize  Subnetwork constructed with $L_0$ penalized regression, multi-source gene expression profiling,  and BIC}
\end{figure}
The 22 genes on the subnetwork were then uploaded onto STRING (http://string-db.org/). STRING is an online database for exploring known and predicted protein-protein interactions (PPI). The interactions include direct (physical) and indirect (functional) associations. The predicted methods for PPI implemented in STRING  include text mining, national databases, experiments, co-expression, co-occurrence, gene fusion, and neighborhood on the chromosome. The PPI network for the 22 genes  are presented in Figure 3.
\begin{figure}[htb!]
\centering
 \includegraphics[scale=0.6]{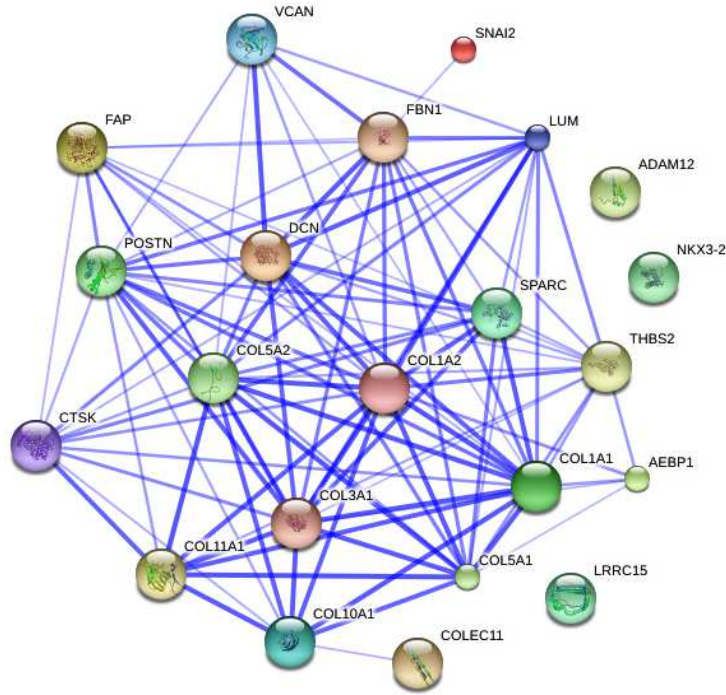}
\caption{\footnotesize  Known and predicted protein protein interactions with the 22 genes on the subnetwork of Figure 2, where nodes represent proteins (genes) and edges indicate the direct (physical) and indirect (functional) associations. Stronger associations are represented by thicker lines.}
\end{figure}
Comparing Figure 3 and Figure 2,  We conclude that the 22  identified genes on the subnetwork of Figure 2 are functioning together and have enriched important biological interactions and associations. Ninteen out of 22 genes on the survival associated subnetwork also have interactions on the known and predicted PPI network, except for genes LRRC15, ADAM12, and NKX3-2. Even though they are not completely identical, many interactions on our subnetwork can also be verified on the PPI interaction network  of Figure 3.  For instance, collagen COL5A2 is the most important genes with the  largest  number of degrees (7) on our subnetwork. Six out of 7 genes that link to COL5A2 also have direct edges on the PPI network.  Those direct connected genes  (proteins) include FAP, CTSK, VCAN, COL1A1, COL5A1, and COL11A1. The remaining gene SNAI2 was  indirectly linked  to COL5A2 through FBN1 on the PPI network. In addition, one of the other important genes with the degree of the node (6) is Decorin (DCN). 4 out of 6 genes directly connected to DCN on  our subnetwork were confirmed on the PPI network, including FBN1, CTSK, LUM, and THBS2. The remain two genes (SNAI2, and COLEC11) are  indirectly connected to DCN on the PPI network.  As indicated on Figure 2, the remaining 5 important genes with  degree of node 4 are POSTN, CTSK, COL1A1, COL5A1, and COL10A1, and 8 genes with degree of node 3 are FBN1, LUM, LRRC15, COL11A1, THBS2, SPARC, COL1A2, and FAP, respectively. Furthermore, those 22 genes are involved in the biological process of GO terms, including extracellular matrix organization and disassembly  and collagen catabolic, fibril, and metabolic processes. They are also involved in  several important KEGG pathways including ECM-receptor interaction, Protein Digestion and Absorption, Amoebiasis, Focal Adhesion, and TGF-beta Signaling pathways. Finally, a large proportion of the 22 genes are known to be  associated with poor overall survival (OS) in ovarian cancer.  For instance, VCAN and POSTN were demonstrated in \textit{vitro} to be involved in ovarian cancer invasion induced by TGF-$\bt$ signaling (Yeung \et, 2013), and COL11A1 was shown to increase continuously during ovarian cancer progression and to be highly over-expressed in recurrent metastases. Knockdown of COL11A1 reduces migration, invasion, and tumor progression in mice (Cheon \et 2014). Other genes such as FAP, CTSK, FBN1, THBS2,  SPARC, and COL1A1 are also known to be ovarian cancer associated (Riester  \et, 2014; Zhao \et, 2011;  Zhang \et, 2013; Gardi \et, 2014; Tang $\&$ Feng 2014; Yu \et, 2014). Those genes contribute to cell migration and the progression of tumors and may be potential therapeutic targets for ovarian cancer.  Further studies with the rest of the genes on the subnetwork are required to explore their biological mechanisms and  potential clinical applications.

\section{Conclusions}
We proposed an efficient  EM algorithm for variable selection with $L_0$ regularized regression. The  proposed algorithm finds the optimal solutions of $L_0$, through solving a sequence of $L_2$ based ridge regressions. Given an initial solution, the algorithm will be guaranteed to converge to a unique solution under mild conditions, and the EM algorithm will be closer to the optimal solution after each iteration. Asymptotic properties, namely consistency and oracle properties are established under mild conditions. Our method apply to fixed, diverging, and ultra-high dimensional problems. We compare the performance of $L_0$ regularized regression and LASSO with simulated low and high dimensional data.   $L_0$ regularized regression outperforms LASSO by a substantial margin under different correlation structures.  Unlike LASSO, which selects more features than necessary, $L_0$ regularized regression chooses the true model with high accuracy, less bias, and smaller test MSE, especially when the  correlation is weak.  Cross-validation with the computation of the entire regularization path is computationally intensive and time consuming. Fortunately $L_0$ regularized regression does not require it. The optimal $\lam_{opt}$ can be directly determined from AIC, BIC, and RIC. Those criteria are optimal under appropriate conditions.  We demonstrate that both AIC and BIC performed well when compared to  cross-validation. Therefore, there is a big computational advantage  of $L_0$, especially with BIGDATA.  In addition, We demonstrate that $L_0$ regularized regression controls the false discovery (positive) rate (FDR) well with both AIC and BIC with the simulation of graphical models. The FDR is very low under different sample sizes with both AIC and BIC. Controlling  FDR is crucial for biomarker discovery and computational biology, because further verifying the candidate biomarkers is time-consuming and costly. We applied our proposed method to construct a network  for ovarian cancer from multi-source gene-expression data, and identified a subnetwork that is important both biologically  and clinically. We demonstrated that we can identify biologically important genes and pathways efficiently. Even though we demonstrated our method with gene expression data, the proposed method can be used for RNA-seq, and metagenomic data, given that the data are appropriately normalized.

\section*{Appendix}
The proposed approach for $L_0$ regularized regression method can be extended to a general $L_p$ $p \in [0, 2]$ naturally. Mathematically, the general $L_p$ problem can be  defined as:
\[ E = \frac{1}{2}||\my - X\ta||^2 + \frac{\lam}{2}\sum_{j=1}^m|\ta|^p, \]
which is equivalent to
\begin{align*}  E  &= \frac{1}{2}||\my -X\ta||^2 + \frac{\lam}{2} \sum_{j \in m} \frac{\ta_j^2}{\eta_j^{2-p}}  \\
               \eta & =\ta.
\end{align*}
Similar ideas in the manuscript can be used to get the the following equation for the general $L_p$EM method:
\[ \eta^{2-p}\odot\frac{\pa E}{\pa \ta}=\lam \ta -\eta^{2-p}\odot X^t(\my - X\ta) = \lam \ta  - X_{\eta}^t(\my - X\ta) = 0,  \]
where $X_{\eta}^t = [\eta^{2-p}\odot\mx_1^t, \ldots, \eta^{2-p}\odot\mx_n^t]$.
Solving Equation (\ref{eq9}),  we have the following explicit solution.
\begin{align*}
      \ta &= (X_{\eta}^tX + \lam I) ^{-1}X_{\eta}^t\my  \\
      \eta &= \ta,
\end{align*}
The general $L_p$EM algorithm is as follows:

\begin{tabular}{l}
$L_p$\textbf{EM Algorithm:}\\\hline
Given a $0<\lam\le \lam_{\max}$,and $p \in [0, 2]$,  small numbers $\epsilon$ and $\varepsilon$,\\
 and training data $\{X, \my\}$,\\
Initializing $\ta =(X^tX +\lam I)^{-1}X^t\my$, \\
While 1,\\
 \;\;\;\;\; E-step:   $ \mf{\eta} = \ta $  \\
\;\;\;\;\;  M-step: $X_{\eta}^t = \eta^{2-p}\odot X^t = [\eta^{2-p}\odot\mx_1 ,\ldots, \eta^{2-p}\odot \mx^t_n]$\\
\;\;\;\;\;\;\;\;\;\;\;\;\;\;\;\text{ }\text{ } \text{ }$ \ta = (X_{\eta}^tX + \lam I) ^{-1}X_{\eta}^t\my $\\
\;\;\;\;\;\;if $||\ta -\eta|| < \varepsilon$, Break;  End\\
End \\
 $\ta(|\ta| <\epsilon) = 0$. \\ \hline\\
\end{tabular}

\subsection*{References}

\begin{description} %\itemsep=-\parsep \itemindent=-1.3cm

\item[ ]Akaike, H. (1974). A new look at the statistical model identification. IEEE T. Automat. Contr. 19, 716–723.

\item[ ]Bunea, F., Tsybakov, A, and Wegkamp, M. (2007). Sparsity oracle inequalities for the Lasso.
{\sl Electron. J. Stat.}, {\bf 1}, 169-194.

\item[ ]Candes, E.J. and Plan, Y. (2009). Near-ideal model selection by $l_1$ minimization. {\sl Ann. Statist.}, {\bf 37}(5A), 2145-2177.

\item[ ]Cheon DJ, Tong Y, Sim MS, Dering J, Berel D, Cui X, Lester J, Beach JA, Tighiouart M, Walts AE, Karlan BY, Orsulic S. (2014), A collagen-remodeling gene signature regulated by TGF-β signaling is associated with metastasis and poor survival in serous ovarian cancer. Clin Cancer Res. 2014 Feb 1;20(3):711-23. doi: 10.1158/1078-0432.

\item[ ]Chi Z. (2009). $L_0$ regularized estimation for nonlinear models that have sparse underlying linear structures. arXiv:0910.2517v1 [math.ST] 14 Oct 2009.

\item[ ]Dicker, L., Huang, B. and Lin, X. (2012). Variable selection and estimation with the seamless-$L_0$
 penalty. Statistica Sinica. In press. doi: 10.5705/ss.2011.074.

\item[ ]Fan, J. and Li, R. (2001). Variable selection via nonconcave penalized likelihood and its oracle properties. J.
Am. Stat. Assoc. 96, 1348–1361.

\item[ ]Foster, D. and George, E. (1994). The risk inflation criterion for multiple regression. Ann. Statist. 22, 1947–1975.

\item[ ]Gardi NL, Deshpande TU, Kamble SC, Budhe SR, Bapat SA. (2014), Discrete molecular classes of ovarian cancer suggestive of unique mechanisms of transformation and metastases. Clin Cancer Res. 2014 Jan 1;20(1):87-99.

\item[ ]Gyorffy B, L$\acute{a}$nczky A, Sz$\acute{a}$ll$\acute{a}$si Z. (2012), Implementing an online tool for genome-wide validation of survival-associated biomarkers in ovarian-cancer using microarray data from 1287 patients. Endocr Relat Cancer. 2012 Apr 10;19(2):197-208. doi: 10.1530/ERC-11-0329.

\item[ ]Lee HK, Hsu AK, Sajdak J, Qin J, Pavlidis P (2004), Coexpression analysis of human genes across many microarray data sets. Genome Res. 2004 Jun;14(6):1085-94.

\item[ ] Lin, D., Foster, D. P., $\&$ Ungar, L. H. (2010). A risk ratio comparison of l0 and l1 penalized regressions. University of Pennsylvania, Tech. Rep.

\item[ ]Liu H, Roeder K, and Wasserman L. (2010) Stability approach to regularization selection for high dimensional graphical models. Advances in Neural Information Processing Systems, 2010.

\item[ ]Liu Q, Ihler A  (2011),  Learning scale free networks by reweighted $L_1$ regularization. AISTATS (2011).

\item[ ] Liu Y, Wu Y (2007), Variable Selection via A Combination of the $L_0$ and $L_1$ Penalties, Journal of Computational and Graphical Statistics, 16 (4): 782–798.

\item[ ]Liu Z, Lin S, Tan M. (2010) Sparse support vector machines with Lp penalty for biomarker identification. IEEE/ACM Trans Comput Biol Bioinform. 7(1): 100-7.

\item[ ]Lu  H and Zhang  Y  (2013), Sparse Approximation via Penalty Decomposition Methods, SIAM Journal on Optimization, 23(4):2448-2478.

\item[ ]Mancera L and Portilla  J. $L_0$ norm based Sparse Representation through Alternative Projections. In Proc. ICIP, 2006

\item[ ]Mazumder R, Friedman, JH,  Hastie  T  (2011), SparseNet : Coordinate Descent with Non-Convex Penalties, JASA,  2011.

\item[ ]Meinshausen, N. \& Bühlmann, P. Stability selection. J. R. Stat. Soc. Series B Stat. Methodol. 72, 417–473 (2010).

\item[ ]Peng,J, Wang P, Zhou N, and Zhu  J. (2009), Partial correlation estimation by joint sparse regression models. JASA, 104 (486):735-746, 2009.

\item[ ]Riester M, Wei W, Waldron L, Culhane AC, Trippa L, Oliva E, Kim SH, Michor F, Huttenhower C, Parmigiani G, Birrer MJ. (2014), Risk prediction for late-stage ovarian cancer by meta-analysis of 1525 patient samples.
    J Natl Cancer Inst. 2014 Apr 3;106(5). pii: dju048. doi: 10.1093/jnci/dju048.

\item[ ]Schwarz, G. (1978). Estimating the dimension of a model. Ann. Statist. 6, 461–464.

\item[ ]Tang L, Feng J. (2014), SPARC in Tumor Pathophysiology and as a Potential Therapeutic Target. Curr Pharm Des. 2014 Jun 19. [Epub ahead of print].

\item[ ]Tibshirani, R. (1996). Regression shrinkage and selection via the lasso. J. Roy. Stat. Soc. B. 58, 267–288.

\item[ ]Yeung TL, Leung CS, Wong KK, Samimi G, Thompson MS, Liu J, et al. TGF-beta modulates ovarian cancer invasion by upregulating CAF-derived versican in the tumor microenvironment. Cancer Res 2013;73:5016–28.

\item[ ]Yu PN, Yan MD, Lai HC, Huang RL, Chou YC, Lin WC, Yeh LT, Lin YW. (2014), Downregulation of miR-29 contributes to cisplatin resistance of ovarian cancer cells. Int J Cancer. 2014 Feb 1;134(3):542-51.

\item[ ]Zhang W, Ota T, Shridhar V, Chien J, Wu B, Kuang R. (2013), Network-based survival analysis reveals subnetwork signatures for predicting outcomes of ovarian cancer treatment. PLoS Comput Biol. 2013;9(3):e1002975. doi: 10.1371/journal.pcbi.1002975.

\item[ ]Zhao G, Chen J, Deng Y, Gao F, Zhu J, Feng Z, Lv X, Zhao Z (2011), Identification of NDRG1-regulated genes associated with invasive potential in cervical and ovarian cancer cells.  Biochem Biophys Res Commun. 2011 Apr 29;408(1):154-9.

\item[ ]Zou, H. (2006). The adaptive lasso and its oracle properties. J. Am. Stat. Assoc. 101, 1418–1429.

\item[ ]Zou, H. and Zhang, H. (2009). On the adaptive elastic-net with a diverging number of parameters. Ann.
Statist. 37, 1733–1751.

\item[ ]Zhang, C. (2010). Nearly unbiased variable selection under minimax concave penalty. Ann. Statist. 38,
894–942.

\end{description}

\end{document}